# Meta-neural-network for Realtime and Passive Deep-learning-based Object Recognition

Jingkai Weng[1*], Yujiang Ding[1*], Chengbo Hu[1], Xue-feng Zhu[2], Bin Liang[1†], Jing Yang[1] and Jianchun Cheng[1†]

[1]*Key Laboratory of Modern Acoustics, MOE, Institute of Acoustics, Department of Physics, Collaborative Innovation Center of Advanced Microstructures, Nanjing University, Nanjing* 210093*, P. R. China*

[2]*School of Physics and Innovation Institute, Huazhong University of Science and Technology, Wuhan, Hubei 430074, P. R. China*

[*]These two authors contributed equally to this work.

[†]Correspondence and requests for materials should be addressed to B.L. (email: liangbin@nju.edu.cn) or to J.C. (email: jccheng@nju.edu.cn)

**Abstract**

Deep-learning recently show great success across disciplines yet conventionally require time-consuming computer processing or bulky-sized diffractive elements. Here we theoretically propose and experimentally demonstrate a purely-passive "meta-neural-network" with compactness and high-resolution for real-time recognizing complicated objects by analyzing acoustic scattering. We prove our meta-neural-network mimics standard neural network despite its small footprint, thanks to unique capability of its metamaterial unit cells, dubbed "meta-neurons", to produce deep-subwavelength-distribution of discrete phase shift as learnable parameters during training. The resulting device exhibits the "intelligence" to perform desired tasks with potential to address the current trade-off between reducing device's size, cost and energy consumption and increasing recognition speed and accuracy, showcased by an example of handwritten digit recognition. Our mechanism opens the route to new metamaterial-based deep-learning paradigms and enable conceptual devices such as smart transducers automatically analyzing signals, with far-reaching implications for acoustics, optics and related fields.

It is a fundamental problem in wave physics to detect and recognize the geometric shapes of objects by properly analyzing the scattered wave, representing the most basic challenge behind a plethora of important applications. Representatively, in acoustics, typical examples range from medical ultrasound imaging[1] to industrial non-destructive evaluation[2,3] to underwater detection[4]. In contrast to the conventional mechanisms that rely on human experts such as physicians interpreting the medical ultrasonic images in the clinic, which would inevitably suffer from low efficiency, potential fatigue and wide variations in pathology[5-7], the recent emergence of computer-assisted deep learning techniques[8] has achieved state-of-the-art performance in the important problem of identification and classification of medical images of scattered acoustic fields such as for detection of anatomical structures and disease diagnosis and so on[7,9,10], among other fascinating applications in speech recognition[11-13], emotion analysis[14-17], etc. In spite of the remarkable improvement in performance and simplification in process, however, such a shift of the burden from human to computers would still arouse the issue of computational complexity, energy supply, device size and cost, owing to their dependence on precise acoustic images that need to be measured via sensor-scanning and computer-based postprocessing. It is therefore essential to continuously pursue new deep-learning-based mechanisms with simpler design, smaller footprint, faster speed and reduced energy consumption and less sensors, which would be vital for the real-world application in many diverse scenarios such as medical imaging where fast and easy assessment of tissues are highly desired.

In this article, we break through such fundamental barriers by introducing a physical mechanism to use a passive "meta-neural-network" comprising a three-dimensional matrix of metamaterial unit cells, with each serving as a "meta-neuron", to mimic an analogous neural network for classical waves with compactness, simplicity and pure-hardware task-solving capability. The recent rapid expansion of the research fields of photonic/phononic crystals[18-22] and metamaterials[23,24] enables unconventional manipulation of wave fields, such as anomalous refraction/reflection[25,26], invisibility cloak[27,28], rectification[29,30], etc., in a deterministic manner, relying on rational design based on human knowledges. The past few years witness considerable efforts devoted to applying machine learning in these artificial structures, but merely aiming at designs of active imaging devices with reduced complexity[31] or metamaterials for producing specific wave fields[32-35]. Recently passive neural networks are proven possible by using diffractive layers with locally-modulated thickness according to machine-learning training results[36], which generates quasi-continuous phase profiles and results in significant phase variation only over wavelength-scale distance. In contrast, the mechanism proposed here represents the first attempt to empower passive metamaterials, with the extraordinary capability to provide abrupt phase shift within deep-subwavelength scales in all three dimensions, with the "intelligence" to perform complex machine learning tasks. We prove that such capability is pivotal for the equivalence between conventional and the proposed neural network and use a computer to train the designed meta-neural-network by iteratively adjusting the whole phase profile of each 2D layer of meta-neurons. The resulting meta-neural-network features planar profile, high spatial density of meta-neurons and subwavelength thickness of each meta-neural-layer, which are particularly crucial for acoustic waves that generally have macroscopic wavelength. More importantly, the simplicity, compactness and efficiency of meta-neural-network enables accurate recognition of objects containing tiny details and complex patterns with compact device orders of magnitudes smaller than diffractive components-based designs in a totally passive, real-time, sensor-scanning-free and postprocessing-free manner, as will be demonstrated hereafter.

Figure 1 schematically shows our proposed mechanism of constructing an acoustic meta-neural-network comprising multiple parallel layers of subwavelength meta-neurons for passive and real-time recognition and classification of objects by the geometric shape. The object to be examined is illuminated normally by a monochromatic plane wave, and the meta-neural-network is located at the transmitted side to receive the scattered acoustic wave produced by the object. The key role of the meta-neural-network is to interact with the incident wave after it is rebounded by the object and thereby converge the acoustic energy, which would scatter into all different directions in its absence, to the desired region on a detection plane behind the last layer instead of being scattered into all directions in the absence of this device, as illustrated in Fig. 1(**a**). For explaining the recognition criterion of meta-neural-network, we exemplify the detection plane for a typical case where 10 handwritten digits, from 0 to 9, are chosen as the object for recognition. The detection plane includes 10 identical square regions assigned respectively for these 10 objects (Fig. 1(**b**)). For a specific object, only when the output signal eventually yielded by the meta-work is accurately redistributed on the detection plane such that the total intensity in the expected region corresponding to this digit is higher than the rest regions, can the recognition and classification be considered successful. For better mimicking the real-world applications, here we do not directly translate the image recognition mechanism for visible light to acoustics by simply using the image of digits as the input pattern and, instead, attempt to realize real-time and high-accuracy recognition of object by appropriately analyzing its scattered wave field. For better mimicking the real-world applications, here we do not directly translate the image recognition mechanism for visible light to acoustics by simply using the image of digits as the input pattern and, instead, attempt to realize real-time and high-accuracy recognition of object by appropriately analyzing its scattered wave field.

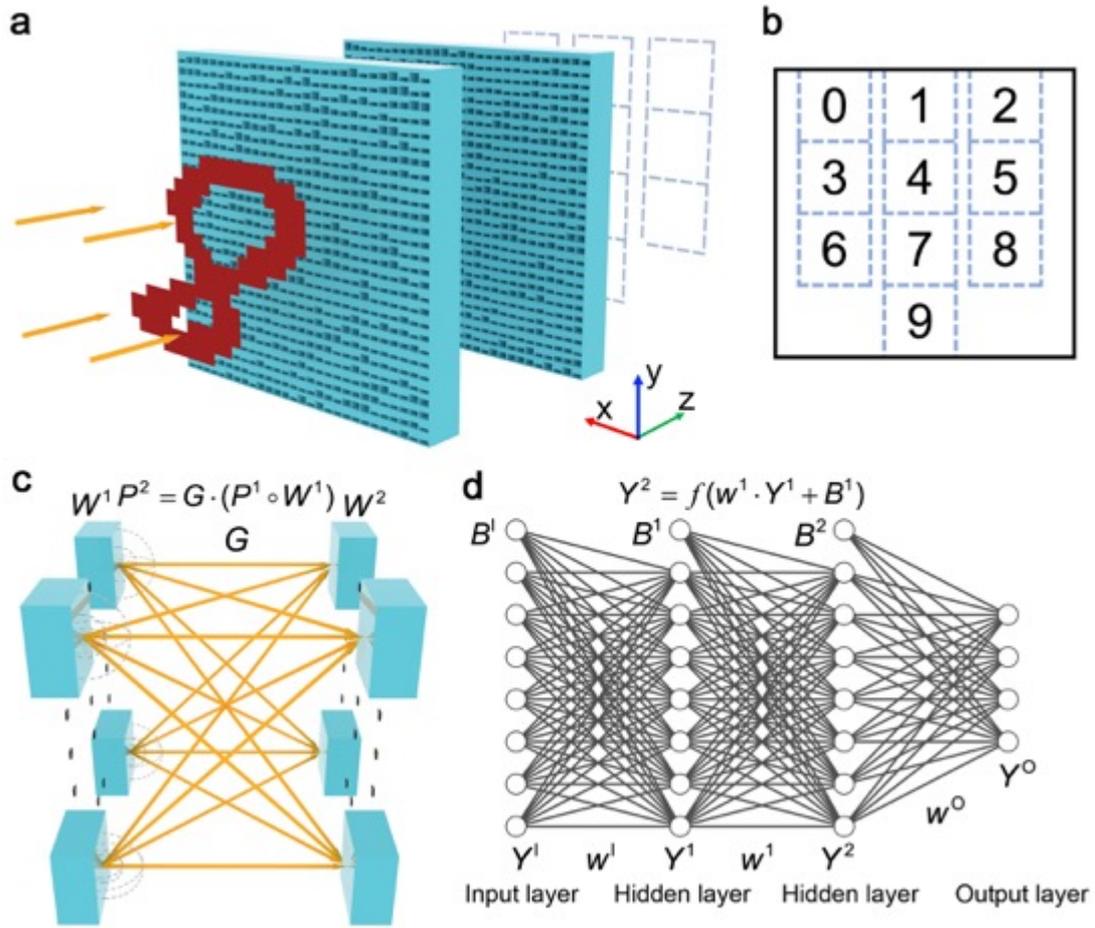

**Figure 1 | Schematic of the proposed mechanism for realizing passive and real-time object recognition based on an acoustic meta-neural-network.** **(a)** The proposed meta-neural-network with network parameters given by a computer-aided training process is capable of converging the scattered energy from the object (chosen as handwritten digit "8" here) into the region corresponding to this digit (marked by dot-line boxes behind the last layer). **(b)** shows a 2D view of the detection regions assigned to the ten handwritten digits from zero to nine for recognition and classification. **(c)** schematically illustrates the interaction between two adjacent 2D layers of meta-neurons, which physically corresponds to the wave propagation from each metamaterial unit cell on the 1st layer to all the unit cells on the 2nd one, after undergoing the phase-amplitude modulation by the 1st layer and free-space diffraction in between

(described by $W^1$ and *G* respectively). (**d**) A conventional neural network is illustrated for comparison.

First we consider the propagation of scattered wave in such a multi-layered acoustic system. As the fundamental building block of our designed meta-neuron-network, each meta-neuron modulates the amplitude and phase of the incident wave as the wave propagates through it. Then the outgoing wave on the transmitted side serves as second sources and becomes the input signal for the next layer, as governed by Huygens' principle[37].

Thus the relationship between the wave fields on two neighboring layers in our meta-neural-network can be written as

$$P^{l+1} = G^l \cdot (P^l \circ W^l) \qquad (1)$$

where vector $P^l$ denotes the input wave of the $l$-th layer of meta-neurons, $G^l$ is the wave propagation matrix (see the Supplementary Note 1 and 2), $W^l = t^l \exp(j\varphi^l)$ is the modulation introduced by the meta-neurons on the $l$-th layer with $t^l$ and $\varphi^l$ referring to the amplitude and phase modulation respectively, "∘" denotes the element-wise multiplication. While the conventional neural network can be written as

$$Y^{l+1} = f(w^l \cdot Y^l + B^l), \qquad (2)$$

where $f$ is the nonlinear active function, $w^l$ is the weight and $B^l$ is bias.

Comparison of Eqs. (1) and (2) clearly reveals the equivalence between our proposed meta-network and a conventional neural network. To be specific, the modulation of each meta-neuron on the transmitted wave mimics the "bias", and the wave propagation function between two neighboring layers of meta-neurons serves as the nonlinear active function. As a result of such equivalence, one can strictly prove that in our meta-neural-network, each meta-neuron connects to all the meta-neurons on the neighboring layer. Given that the transmission loss is trivial for all the meta-neurons, the output wave of an individual meta-neuron is solely determined by the shift in propagation phase it provides for the input wave, and the phase modulation essentially plays the same role as the weight in conventional deep-neural-network. Hence the phase shifts of meta-neurons are chosen as the learnable parameters for training in the machine learning process as we will show later. Due to deep-subwavelength size of individual meta-neurons, each single 2D layer of meta-neurons can be alternatively treated as acoustic metamaterial composed of a monolayer of subwavelength unit cells with vanishing thickness in terms of wavelength[38-40]. Apparently, only by going beyond the conventional limit in natural material that propagation phase can only gradually accumulates along the path by utilizing the unique capability of metamaterial to produce abrupt phase discontinuity within subwavelength dimension can such an assumption of, meta-neural-network model with compact size and ultra-fine phase resolution be considered valid[41-45].

Notice that the proposed strategy needs no measurement of the original scattered field nor reconstruction of the precise acoustic image, exempt from the burden on the cost and time in conventional computer-assisted deep learning paradigms which will further increase when the object complexity is enhanced or the detection region is enlarged. Limited by the current technology, this will result in many challenges including implementing large-scaled phased arrays[46], fabricating subwavelength sensor (e.g. piezoelectric transducer) and accelerating measurements and analysis of huge amount of sound field data. In stark contrast, the meta-neural-network performs detection and computation simultaneously due to the parallel interaction between wave and meta-neurons without sensor-scanning or postprocessing, which accomplishes once the incident wave passes regardless of the resolution or number of meta-neurons, and the output field only needs to be measured at the receiving end with fixed number of sensors (as shown in Fig. 1(**a**)) as few as the possible classification types of objects (chosen as 10 here), no matter how complicated the target is. In additional to these advantages of passive elements in terms of speed and simplicity, our proposed meta-neural-network with compact planar geometry and ultra-fine phase resolution enables downsizing the device to the scale unattainable with diffractive components.

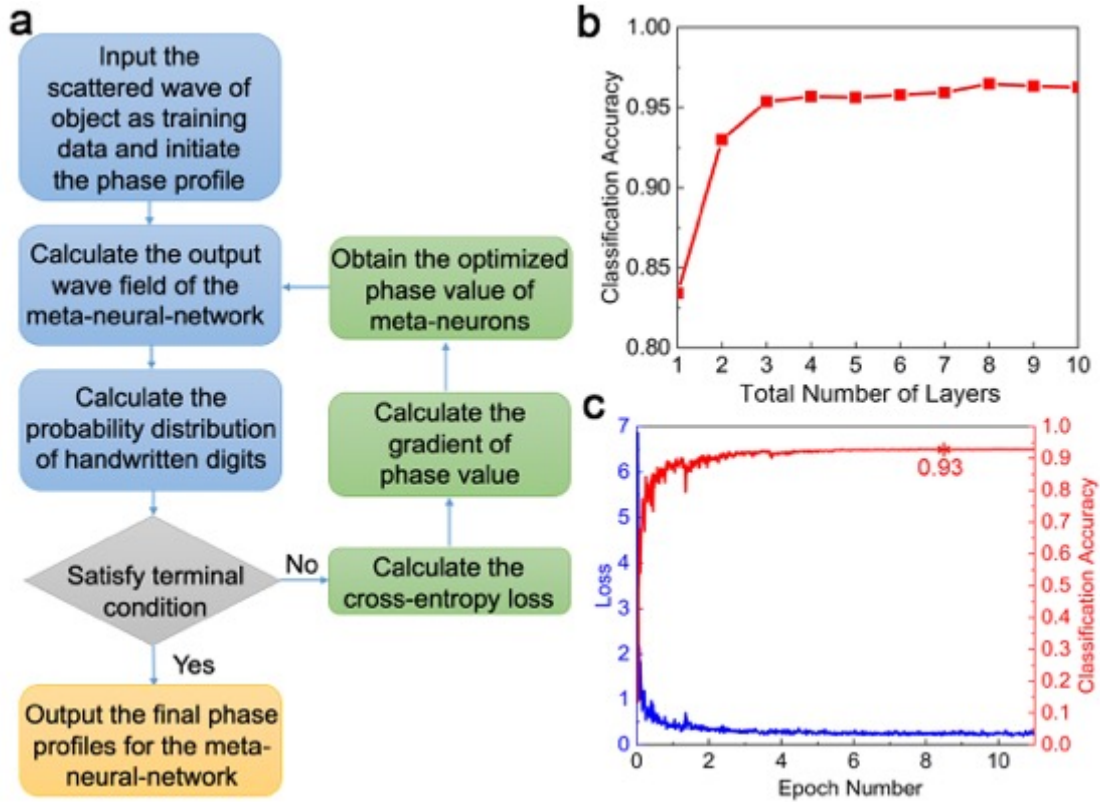

**Figure 2 | Simulated results for deep learning training of the meta-neural-network.** **(a)** The chart flow of the training process that uses the scattered wave produced by different objects as training data and calculates the output of meta-neural-work to iteratively tune the phase value of each meta-neuron, until achieving the maximal probability of converging the scattered energy produced by a specific class of object into the predesigned region. **(b)** shows the classification accuracy increases as the total number of layers adds from one to ten. **(c)** depicts the simulated dependence of loss value and classification accuracy on the epoch number, showing that the accuracy increases with epoch number and eventually reaches the maximum (93%) in the training process of our designed meta-neural-network.

To manifest the unique advantages of our proposed meta-neural-networks with compactness and efficiency, we first choose to demonstrate via both simulation and experiment the recognition of MNIST (Modified National Institute of Standards and Technology) handwritten digits on a scale approximately one order of magnitude smaller than attainable with deep-learning-based diffractive layers. The database contains 55000 training images, 5000 validation images, and 10000 testing images. For simplifying the design and fabrication of meta-neural-network sample in the following experiments, we avoid simultaneous adjustment of amplitude and phase for the transmitted wave and only use phase modulation with the transmission efficiency being set to be 1, which does not appreciably affect the accuracy of the resulting device as we demonstrate via numerical simulation (see the Supplementary Note 3). Each object is implemented based on a binary image formed by rounding up the grayscale value of each pixel in the corresponding MNIST image (see the Supplementary Note 4). The details of training process are shown in Fig. 2(**a**), we use the simulated scattered waves produced by the objects, instead of the object images themselves usually used for image recognition purpose, as the training data and thereby train the meta-neural-network to extract the typical scattering pattern of each class of objects. The cross-entropy loss function[47] which is commonly used in classification problem is introduced (see the Supplementary Note 1), and the gradient of phase value is calculated through error back-propagation algorithm[48]. We adjust the phase values of meta-neurons in search of the minimum of loss value corresponding to the maximum likelihood of making the total acoustic intensity in the target region higher than the others for as many digits as possible in the MNIST database. By iteratively feeding training data, the classification accuracy of testing data keeps increasing and eventually becomes stable within 6 epochs.

In our simulation, the operating frequency is set to be 3kHz (corresponding to a wavelength of approximately 11.4 cm in air) such that the experimental sample of meta-neural-network is of moderate size which facilitates both the 3D printing fabrication of subwavelength meta-neurons and the sound field measurement in anechoic chamber. As a specific design, each layer is chosen to consist of $28 \times 28$ (784 in total) meta-neurons, equal to the number of pixels in a handwritten digits picture in the MNIST database. Each individual meta-neuron is assumed to have a sub-wavelength size in each dimension, consistent with the actual size of the practical metamaterial we will implement in the measurement. Specifically, the transversal size of the meta-neuron is 2 cm (smaller than 1/5 wavelength), which helps to ensure deep-subwavelength resolution of meta-neural-network that is vital for the high-accuracy recognition for more sophisticated cases. The axial distance between two neighboring layers is set to be 17.5 cm. After its training, the design of our meta-neuron digit classifier was numerically tested by 10000 images from MNIST test dataset.

Here we choose a design of meta-neural-network consisting of two layers of metamaterial only for a balance between the classification accuracy and efficiency, based on the our numerical analysis on the dependence of accuracy on the layer number as shown in Fig. 2(**b**) which indicates that the increase rate of accuracy with respect to layer number becomes much slower for designs containing more than two layers. The accuracy of recognition by such a simple bilayer structure can reach 93%, which is considerably high given the significant acceleration of training process, reduction of meta-neuron number and downscaling of resulting device. By further increasing the total number and spatial resolution of meta-neurons contained in our meta-neural-network, it is still possible to achieve higher accuracy of recognition, at the cost of enlarging the whole meta-neural-network and enhancing the fabrication precision of unit cells. Figure 2(**c**) shows that the loss value keep decreasing while classification accuracy keep increasing before a maximum of 93% is reached.

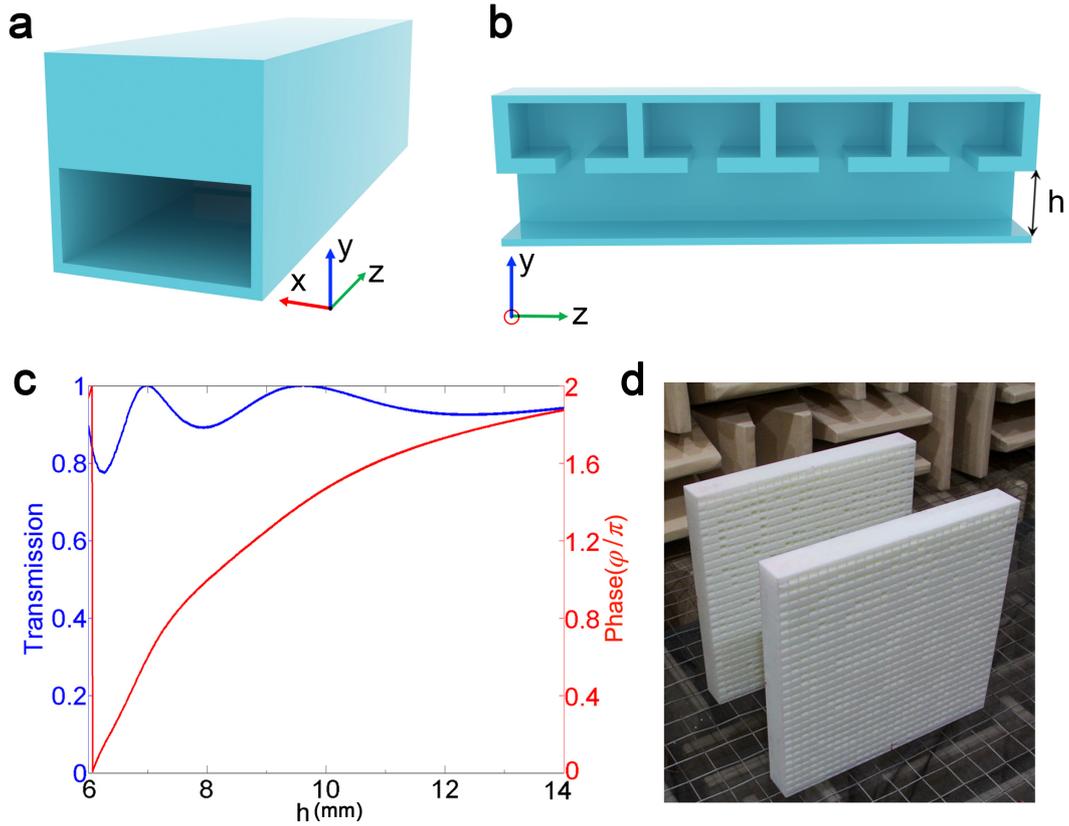

**Figure 3 | Implementation of meta-neuron as basic building block of the proposed acoustic meta-neural-network.** (**a**) 3D schematic view of an individual meta-neuron which is a metamaterial unit cell with a hybrid structure. (**b**) shows the 2D cross-section view of the meta-neuron illustrated in (A) which is formed by coupling four local resonators and a straight pipe for producing a coupled resonance that compensates the impedance mismatch in the phase modulation. (**c**) The simulated phase discontinuity provided by this unit cell as a function of the height of pipe, h, shows that adjustment of this single parameter ensures full 2π phase control while keeping near-unity transmission efficiency. (**d**) The photo of the meta-neural-network prototype fabricated via 3D printing technique, which are composed two layers of meta-neurons implemented by metamaterial unit cells with subwavelength size in all dimensions.

Next we perform experimental measurements to verify our proposed mechanism. As a practical implementation, in the current study we propose to design a metamaterial unit cell composed of four local resonators and a straight pipe[49], as illustrated in Fig. 3.

Here the series connection of cavities supports strong local resonance for effectively slowing down the propagation of wave as it passes by, enabling free control of the propagation phase within the full 0-to-2π range. On the other hand, the impedance mismatch caused by these local resonators is maximally compensated by the introduction of the straight pipe supporting Fabry-Perot resonance, which results in hybrid resonance and ensures full 2π control of phase shift while keeping a near-unity transmission efficiency. This is verified by the numerical results depicted in Fig. 3(c), which shows that the phase shift can be smoothly tuned within the range from 0 to 2π by adjusting a single structural parameter, $h$ ,which is the height of pipe, with no need of changing the overall size of each subwavelength meta-neuron (2 cm in width and 7 cm in thickness). It is also noticed that such a metamaterial design leaves an average of 50% cross-section open for each layer and keeps the continuity of the background medium, which allows other entities such as light or flow to pass and helps to improve the application of the resulting devices in practical scenarios such as photoacoustic imaging where the opaque acoustic transducers usually block the transmission of light. Based on the parameter dependence of phase shift given by the simulation results shown in Fig. 3(c), we determined the precise geometric parameter for each meta-neuron and fabricated a meta-neural-network comprising two layers with transversal size of 56 × 56 cm.

With our designed meta-neural-network, the handwritten digits in testing dataset have been well classified and the acoustic energy has been redistributed into the target region, as shown in Figs. 4(a,b). In the experiment, we have fabricated 2 sets of steel plates with shapes of handwritten digits (viz., 20 objects in total, and the simulation result is shown in Fig. 4(c)) which are selected from the test images that have been numerically proven capable of being correctly classified by our designed meta-neural-network with each meta-neuron endowed with the ideal phase value given by the computer-aided training process. Good agreement is observed between the theoretical and experimental results as shown in Fig. 4(d) which takes the digit "0" as an example (more details and results in Supplementary Note 5), with both revealing that our designed double layered meta-neural-network accurately redistributes the input energy into the detection region assigned to the object, except for the poor performance of meta-neural-network when recognizing digit "4" which primarily stems from the experimental error (see the Fig. 4(e) and Supplementary Note 6).

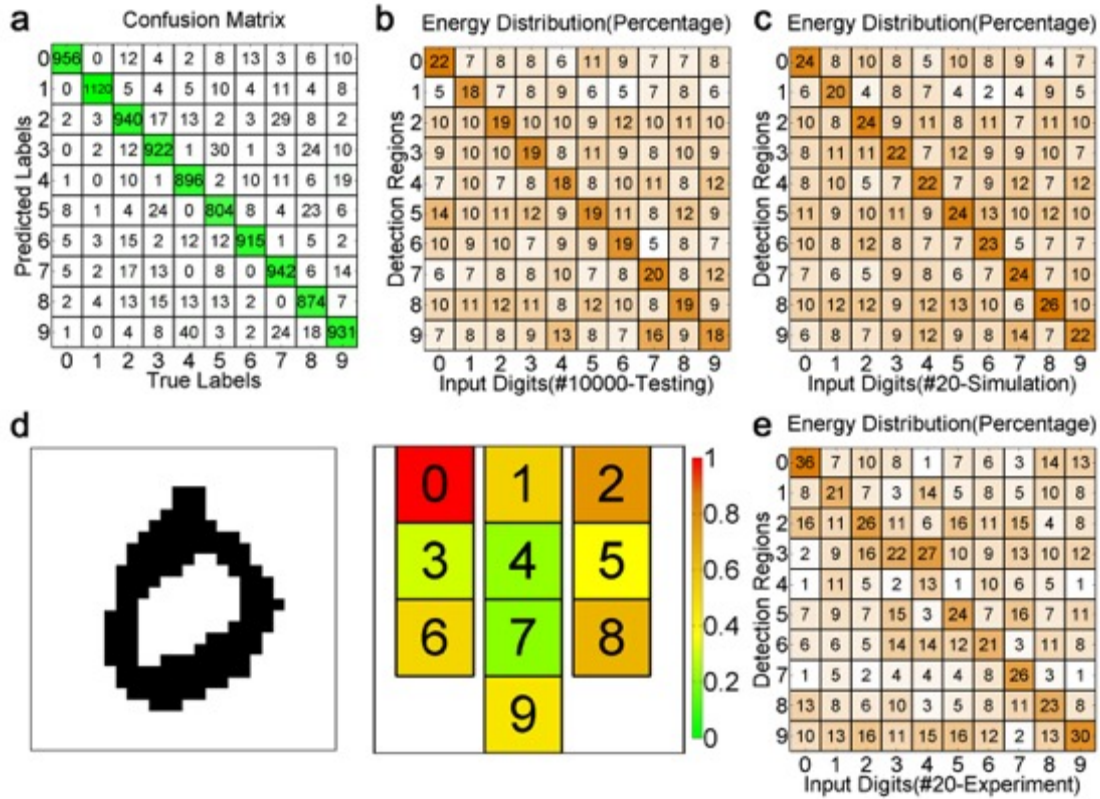

**Figure 4 | Experimental verification of passive and realtime object recognition by acoustic meta-neural-network (a)** shows the confusion matrix for the numerical results of two-layers meta-neural-network with 10000 handwritten digits. **(b)** the energy distribution percentage of 10000 handwritten digits. **(c)** shows the energy distribution of 20 selected digits in simulation. **(d)** total acoustic intensity measured in each detection region corresponding to digits "0". **(e)** is the same as **(c)** but for the experiments.

It is worth mentioning that in the current study we choose to recognize and classify ten handwritten digits which usually acts as the simplest possible example for verifying the effectiveness of deep learning-based techniques. Instead of directly translating the image recognition mechanism for visible light to acoustics by simply using the image of digits as the input pattern, here we demonstrate the possibility of realizing realtime and high-accuracy recognition of object by appropriately analyzing its scattered wave field. This better mimics the real-world applications and suggests the potential of our proposed mechanism to be extended to recognize more complicated objects, e.g., tumors to be diagnosed in ultrasound imaging or defects to be identified in industrial testing, given that sufficiently large database can be accessed for training our meta-neural-network. In addition, the accurate recognition of such objects with bulky size and/or fine structures in terms of wavelength would require high-intensity spatial sampling and large array size, which is demanding for conventional active-device-based approaches but can be readily implemented by our design with unique deep-subwavelength spatial resolution and realtime recognition functionality. With the unprecedented capability of realizing complicated tasks in need of no time delay, energy consumption or postprocessing, our mechanism provides a faster, cheaper and simpler paradigm of deep learning technique, with potential to revolutionize various areas like medical diagnosis and enable novel applications such as realtime assessment of tumors.

In conclusion, we for the first time demonstrate, both theoretically and experimentally, the possibility to unite the ideas of metamaterials and deep-learning to realize a passive and compact meta-neural network performing various complicated tasks such as object recognition via interacting with classical waves. Besides having no dependence on human experts as in computer-based deep learning methods, our proposed meta-neural-network needs no complicated sensor arrays nor high-cost computers, and, in particular, performs real-time recognition without power supply, thanks to its passive nature and parallel wave-interaction, exempt from the heavy burden on the computational hardware in conventional deep learning methods. Furthermore, the meta-neural-networks have small footprint thanks to the subwavelength nature of metamaterials, which is vital for their application in acoustics where acoustic waves generally have macroscopic wavelength but unachievable with diffractive components-based neural networks. Our design with simplicity, compactness and efficiency offers the possibility of miniaturization and integration of deep learning devices, and may even open route to the design of new generation of conceptual acoustic devices such as portable and smart transducers which, as a result of coupling the functionalities of detection and computation, may be able to automatically analyze the backscattered acoustic signals it receives and subsequently complete sophisticated tasks such as evaluating tumors in a totally passive, sensor-scanning-free and postprocessing-free manner. Furthermore, our designed device serves as a new class of "passive deep-learning chips" for power-supply-free yet real-time task-solving purpose, with the ability to inspire relevant researches for other classical waves and to enable novel functional devices such as lightspeed subwavelength meta-neural-networks with significant potential to all-optical on-chip applications.

**Acknowledgements**

This work was supported by the National Key R&D Program of China (Grant No. 2017YFA0303700), the National Natural Science Foundation of China (Grant Nos. 11634006, 11374157 and 81127901), the Innovation Special Zone of National Defense Science and Technology, High-Performance Computing Center of Collaborative Innovation Center of Advanced Microstructures and A Project Funded by the Priority Academic Program Development of Jiangsu Higher Education Institutions. The authors thank Prof. Yu Zhou for fruitful discussions.